\documentclass{article}

\usepackage[preprint,nonatbib]{neurips_2019}

\usepackage[utf8]{inputenc} 
\usepackage[T1]{fontenc}    
\usepackage{hyperref}       
\usepackage{url}            
\usepackage{booktabs}       

\usepackage{graphicx}
\usepackage{amsmath}

\usepackage{dsfont}
\usepackage{multirow}
\usepackage{algorithm}
\usepackage{algorithmic}
\usepackage{bm}
\usepackage{wrapfig}
\usepackage{makecell}
\usepackage{cite}

\title{Nucleus Neural Network: A Data-driven Self-organized Architecture}

\author{
Jia Liu\\
School of Computer Science and Engineering\\
Nanjing University of Science and Technology, Nanjing, 210094, China\\
\texttt{omegaliuj@gmail.com}\\
\And
Maoguo Gong\\
School of Electronic Engineering\\
Xidian University, Xi'an, 710071, China\\
\texttt{gong@ieee.org}\\
\And
Haibo He\\
Department of Electrical, Computer, and Biomedical Engineering\\
University of Rhode Island, Kingston, RI, 02881, USA\\
\texttt{haibohe@uri.edu}\\
}

\begin{document}

\maketitle

\begin{abstract}
Artificial neural networks which are inspired from the learning mechanism of brain have achieved great successes in many problems, especially those with deep layers. In this paper, we propose a nucleus neural network (NNN) and corresponding connecting architecture learning method. In a nucleus, there are no regular layers, i.e., a neuron may connect to all the neurons in the nucleus. This type of architecture gets rid of layer limitation and may lead to more powerful learning capability. It is crucial to determine the connections between them given numerous neurons. Based on the principle that more relevant input and output neuron pair deserves higher connecting density, we propose an efficient architecture learning model for the nucleus. Moreover, we improve the learning method for connecting weights and biases given the optimized architecture. We find that this novel architecture is robust to irrelevant components in test data. So we reconstruct a new dataset based on the MNIST dataset where the types of digital backgrounds in training and test sets are different. Experiments demonstrate that the proposed learner achieves significant improvement over traditional learners on the reconstructed data set.
\end{abstract}

\section{Introduction}
Mimicking the architecture and mechanism of nature creatures' learning for creating efficient and robust learners is one of the challenges in artificial intelligence research \cite{review1}. Artificial neural networks which are inspired form nature neural systems have excellent learning capability in many applications. Specially, neural networks with deep hierarchical layers have achieved many breakthroughs in machine learning, which leads to the great research interests in deep learning \cite{review2}.
Architecture of neural networks indicating the information flow along neurons and connections between them, plays an important role in neural learning. Therefore, architecture optimization has been a research interest for decades with various optimization methods \cite{architecture1,architecture2,architecture3,architecture4,architecture5,architecture6,architecture7,architecture8,architecture9,architecture10}.

\begin{figure}
  \centering
  \includegraphics[width=0.8\linewidth]{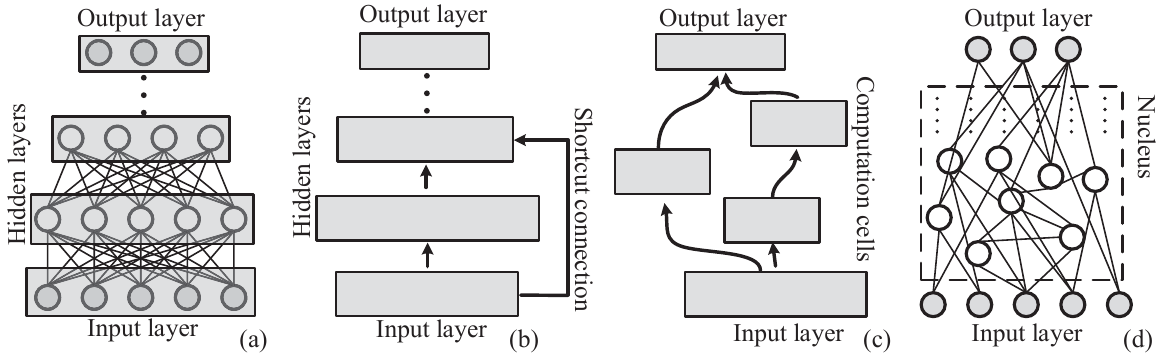}\\
  \caption{Architectures learned by different methods. (a) Multi-layer network (b) Multi-layer network with shortcut connections (c) Architecture learned based on neural network cells (d) Architecture of NNN}\label{fig:architecture}
  \vspace{-10pt}
\end{figure}

Intuitional architecture optimization methods follow the hierarchical architectures, i.e., multiple layers with no connections in the same layer as shown in Fig. \ref{fig:architecture}(a) and find the optimal depth, layer width, kernel size (convolutional neural network), and connections between each two layers. For example, NEAT algorithm used in \cite{architecture2} evolves the nodes, connections and weights for a network. In \cite{architecture9} and \cite{architecture10}, the sparse connecting structure is focused and redundant connections are removed from a hierarchical compact network. It is known that the information process in brain is hierarchical but the architecture in each layer is much more complex than that of artificial neural networks (usually no connections in each layer). There are some attempts trying to add additional skip connections \cite{shortconnections1,shortconnections2,shortconnections3,shortconnections4,shortconnections5} among which the most successful one is the residual network \cite{shortconnections5}.
This motivates the learning of more additional connections in architecture. For example, in \cite{architecture7}, a complex architecture is evolved with involvement of not only multi-layer configurations but also skip connections. Nowadays, there are increasing interests in architecture learning based on network blocks or cells \cite{architecture8,architecture11,architecture12,architecture13}. A block or cell is a directed acyclic graph consisting of an ordered sequence of nodes \cite{architecture8}. The connections between them are optimized to achieve a better performance of classification or other tasks. Such architectures further relax the layer limitation and achieve larger learning capability. In this paper, inspired from the nuclei in brain where architectures are apparently irregular but perfectly process input information, we attempt to fully relax the layer limitation and evolve the network architecture freely given a set of neurons. A neuron is able to connect to any neurons and the connections are organized according to data and tasks. But since we focus on classification problem in this paper, the only limitation is that there are no feed-back connections.
We call the new architecture nucleus neural network (NNN). Fig. \ref{fig:architecture} exhibits the architectures learned by different architecture learning methods.

However, with more free binary variables, searching the optimal connections of NNN is of great difficulty. Most of architecture learning methods search the optimal architecture based on task itself as the objective, for instance, test error for classification. But it is necessary to train the weights and biases to compute the objective. Therefore, it is computationally demanding even though the searching space in them is not that large. For example, 1800 GPU days of reinforcement learning is required to search the optimal architecture in \cite{architecture11} and 3150 GPU days of evolution in \cite{architecture13}. There are also many methods proposed to speed up the learning process, such as weight sharing \cite{architecture14}, Bayesian optimisation \cite{architecture15}, and differentiable network architecture search \cite{architecture8}. However, most of them focus on reducing the searching space or increasing searching efficiency. In NNN, the optimizer should search in a binary space with over ten thousand dimensions which is much larger than that of most existing architecture learning problems. As a consequence, we propose a substitutable but more efficient objective function with respect to the architecture only. We define a connecting density between a neuron and an input neuron. Then if an input neuron is more relevant to an output neuron according to the mutual information computed from the training data, the connecting density between them should be larger. The objective function is defined based on this principle via the modeling approach of products of experts (PoE) \cite{PoE} in order to capture the distribution of observed data directly by the architecture without the involvement of weights and biases. This will greatly reduce the computational complexity of objective function. A binary particle swarm optimizer (BPSO) algorithm \cite{pso1} is utilized to optimize the objective function. After the architecture is optimized, the connecting weights and biases are learned by a novel error driven probability model to well represent the input data.

After training, we find an interesting phenomenon of NNN. Since the architecture is evolved based on the input and output relevance, NNN gives high response to the data where patterns of irrelevant components are never seen by NNN during training. This means that NNN is robust to background changes when trained only by images with pure background in image classification.
To highlight the superiority of NNN, we construct a new data set where the background types of training and test data are different. The potential application is significant but difficult for traditional learners since most of them assume that training and test data are independent but follow identical distribution.

\section{NNN's Architecture Learning}
NNN is composed of input layer, output layer and a nucleus as shown in Fig. \ref{fig:architecture}(d). In the nucleus, there are no regular layers and a neuron can connect to any neurons. But for time independent data, feedback connections are unallowed. The goal of architecture learning is to determine the connections between those neurons in the nucleus.
In order to reduce the computational complexity, we directly model the connections without considering the connecting weights and biases. It is established based on the principle that higher relevance leads to higher connecting density so as to learn the relationship between input and output nodes. First the connecting density is defined.

\subsection{Connecting Density}
To evaluate the architecture, we should consider not only depth but also width of the information flow between an input and an output neurons. Therefore, we define a propagated connecting density with the initial density between an input neuron and itself being 1. Then the density $\mathcal{D}_{ij}(\varphi)$ between a neuron $i$ in the nucleus and an input neuron $j$ can be computed as follows:
\begin{equation}
\label{eq:density}
\mathcal{D}_{ij}(\varphi)=\frac{\sum_{k\in\Omega_i}\mathcal{D}_{kj}(\varphi)}{N_{\Omega_i}}
\end{equation}
where $\varphi$ denotes the architecture of the network which is a binary matrix indicating the connecting status between each pair of neurons with $1$ denoting connected and $0$ unconnected. $\Omega_i$ denotes the set of lower neurons that directly connect to $i$ and $N_{\Omega_i}$ is the number of neurons in the set. It means the average connecting density from the lower neural nodes and can be simply read as the residue of input information. The connecting density starts from the input neurons, propagates through the connections, and finally that between input and output neurons is obtained. The connecting density represents a information path between an input and an output neurons. From Eq. \eqref{eq:density}, higher density means wider and shallower path with more processing nodes and less information degression. If the path is deeper, the information of this input neuron will be disturbed by information from other neurons. Therefore, the density will be lower. The information path should well adapt to the relevance between input and output neurons. Based on this principle, we then construct the model of each output neuron.

\subsection{Modeling Single Output Neuron}
Here we utilize the normalized pointwise mutual information (NPMI) \cite{NPMI} to represent the relevance between input and output neurons for each data. Suppose an input neuron $j$ and an output neuron $i$, given a pair of input and output neurons' status $\{x_j,y_i\}$ with input vector $x$ and referenced output vector $y$, the NPMI between them $\mathds{N}(x_j;y_i)$ can be computed by:
\begin{equation}
\label{eq:NPMI}
\mathds{N}(x_j;y_i)=\frac{\mathds{I}(x_j;y_i)}{\mathds{H}(x_j,y_i)}=\log\frac{P(x_j,y_i)}{P(x_j)P(y_i)}/\log \frac{1}{P(x_j,y_i)}
\end{equation}
where $\mathds{I}$ denotes the pointwise mutual information (PMI) and $\mathds{H}$ denotes the self-information. $P(~)$ denotes the probability which is counted from the dataset. The NPMI normalizes PMI into an interval of $[-1,1]$ with $-1$ denoting the two values will never occur together, $0$ independence, and $1$ complete co-occurrence. Because the probability of a negative valued $\mathds{N}(x_j;y_i)$ is very small and can be omitted, we define the relevance between $x_j$ and $y_i$ as $\mathds{R}(x_j,y_i)=\max(\mathds{N}(x_j;y_i),0)$ \cite{PPMI} to avoid unexpected overflow.

The connecting density should follow the relevance for a better information flow of input and output neurons. Therefore, the possibility that an output neuron and its corresponding information path can well represent the input vector is measured by the cosine distance between relevance and connecting density:
\begin{equation}
\label{eq:experts}
p_i(x,y_i;\varphi)=\cos(\mathcal{D}_i(\varphi),\mathds{R}(y_i))
\end{equation}
where $\mathcal{D}_i(\varphi)$ and $\mathds{R}(y_i)$ respectively denotes the connecting density and relevance vectors with $n$ dimensions, i.e., $\mathcal{D}_i(\varphi)=[\mathcal{D}_{i1}(\varphi),\mathcal{D}_{i2}(\varphi),...,\mathcal{D}_{in}(\varphi)]^T$ and $\mathds{R}(y_i)=[\mathds{R}(x_1,y_i),\mathds{R}(x_2,y_i),...,\mathds{R}(x_n,y_i)]^T$ where $n$ is the number of input neurons. For an architecture that best captures the relevance between an output neuron and input neurons, the possibility, i.e., cosine distance will be close to 1. While if there is no enough connecting density for transforming the information, the possibility will be close to 0.

\subsection{Modeling the Whole Network}
The whole network can be modeled by combining the models of output neurons. For better modeling the high-dimensional space of input data, in this paper, we combine the neurons by multiplying them together and renormalizing as in PoE. PoE has the advantage that they can produce much sharper distributions \cite{PoE}. In NNN, the whole network model is formulated as follows:
\begin{equation}
\label{eq:PoE}
p(x,y;\varphi)=\frac{\prod_ip_i(x,y_i;\varphi)}{\sum_{(\chi,\gamma)}\prod_ip_i(\chi,\gamma_i;\varphi)}
\end{equation}
where $(\chi,\gamma)$ denotes one of all the possible data in the whole data space which $(x,y)$ belongs to. The denominator is the renormalization term used to guarantee that the probability sum over the whole data space is 1. The model follows PoE where simple models $p_i(x,y_i;\varphi)$ are multiplied and renormalized to construct a complex model $p(x,y;\varphi)$. Optimizing this model means to increasing the representation capability of observed data while decreasing that of all the other data. Then the architecture $\varphi$ will well capture the distribution of observed data.

However, optimizing this model is of great difficulty due to the unreachable fantasy data in the whole data space. In PoE, the model is optimized by using the gradient of a log-likelihood. Gibbs sampling is used to estimate the gradient expectation of the whole data space. However, in this model, the decision variable is the architecture $\varphi$ which is binary. Such a problem is suitable to be solved by evolutionary algorithms or population based methods \cite{evoldnn1,evoldnn2}. Therefore, we intend to optimize the architecture via a binary particle swarm optimization (BPSO) algorithm. A computable objective function is necessary to be derived.

\subsection{Objective Function}
Fortunately, the value of the denominator in Eq. \eqref{eq:PoE} depends only on the architecture $\varphi$. Optimizing this model amounts to maximizing the numerator wile minimizing the denominator. Therefore, the denominator can be replaced by a computable function with respect to $\varphi$ and taken as a new term added to the numerator. Here we use the L2-norm of connecting density vector of each output neuron:
\begin{equation}
\label{eq:replaced}
\rho(\varphi)=\sum_i\|\mathcal{D}_i(\varphi)\|_2
\end{equation}
The denominator in Eq. \eqref{eq:PoE} aims to represent the whole data space by connecting density. Since the whole data space contains all the possible cases, connecting density controls the representation ability of the architecture for all the possible data. Therefore, minimizing $\rho(\varphi)$ will decrease the denominator in Eq. \eqref{eq:PoE}.

Then the new objective function can be reconstructed by combining the two terms, i.e., numerator in Eq. \eqref{eq:PoE} and Eq. \eqref{eq:replaced}:
\begin{equation}
\label{eq:obj}
\max J(\varphi)=\sum_{(x,y)\in \mathbf{D}}\prod_ip_i(x,y_i;\varphi)-\lambda\rho(\varphi)
\end{equation}
where $\mathbf{D}$ is the training set and $\lambda$ is a user defined parameter that controls the importance of the two terms. In this objective function, the denominator in Eq. \eqref{eq:PoE} is replaced by a simplified function but can achieve similar effectiveness to that of Eq. \eqref{eq:PoE}. Maximizing this function amounts to assigning more probability to the observed data (increasing numerator) while restraining the probability of all the other data (decreasing denominator). Then the architecture will capture the distribution of observed data and well represent the input and output relationship of the data in data set $\mathbf{D}$. Finally we utilize a BPSO algorithm to optimize the objective function.

\subsection{Toy Example of Learned Architecture}
\begin{wrapfigure}[11]{r}{0.5\linewidth}
  \centering
  \includegraphics[width=1\linewidth]{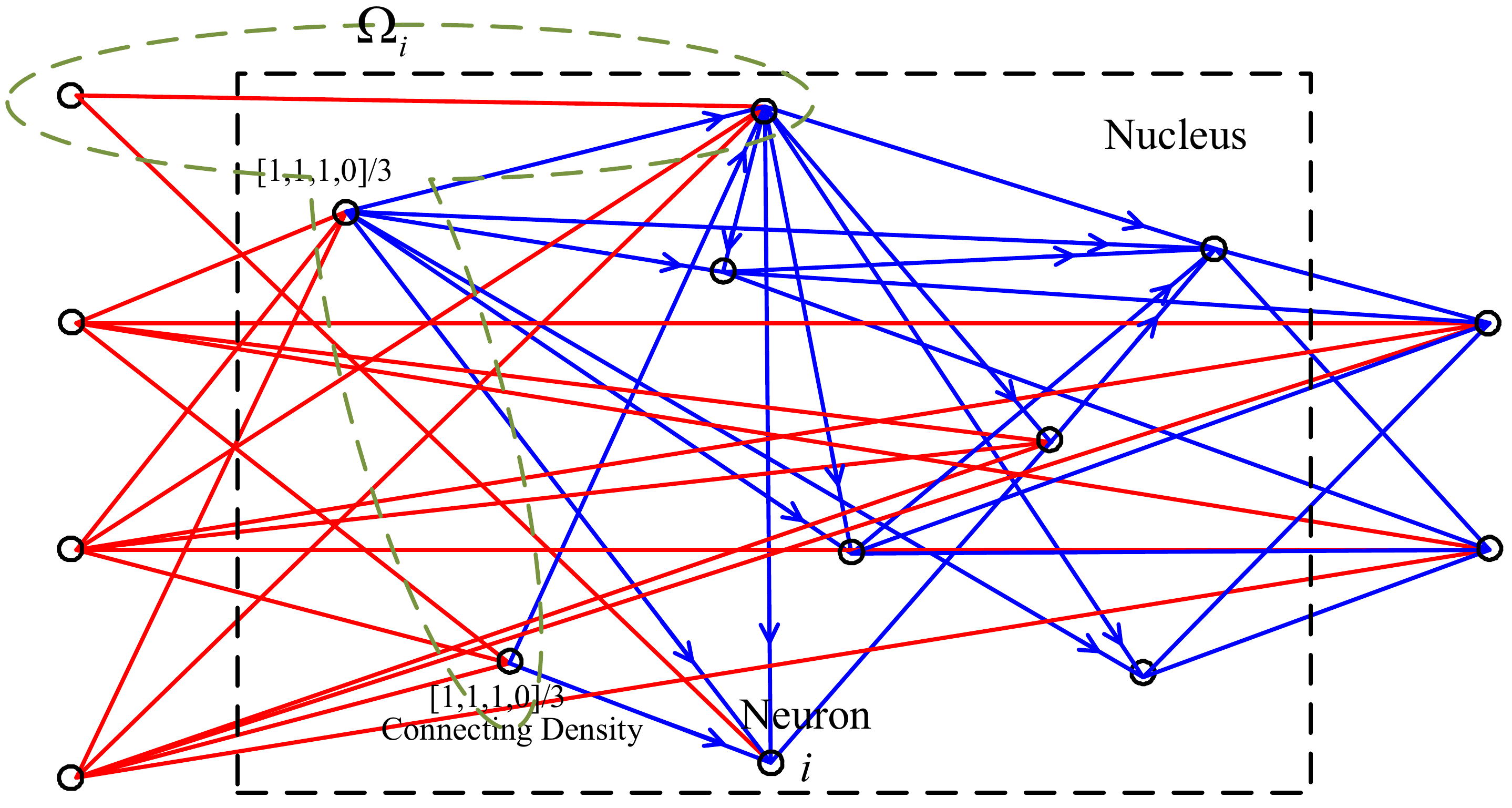}
  \vspace{-15pt}
  \caption{A toy example of learned architecture from a simulated data set. Red lines represents the connections from input neurons.}\label{fig:toyexample}
\end{wrapfigure}

To better explain the whole story, we provide a toy example of the learned architecture from a simulated data set. The data set is constructed as a classification problem. There are 4 attributes in the input data with the first 3 attributes following the joint Gaussian distribution and the last attribute follows the uniform distribution. In the learning process, we set 9 neurons in the nucleus and Fig. \ref{fig:toyexample} shows the learned architecture. It is intuitive that the first 3 attributes contribute more to the classification, and therefore more connections are connected to them. They also directly connect to the output neurons. The last attribute has no contributes to the output neurons, the depth of them is over 3 layers. The learned architecture follows the principle in appearance which demonstrates the effect of the objective function and learning method.

The connecting density of some neurons and $\Omega_i$ of a neuron $i$ are shown Fig. \ref{fig:toyexample}. The initial connecting density of input neurons are 1. As shown in the figure, the connecting density between a neuron and input neurons is represented as a vector. Then they propagate through the whole network and $\mathcal{D}_i(\varphi)$ for each output neuron is obtained.

\section{NNN's Parameter Learning}
After the architecture is learned, the network parameters, i.e., the weights on each connection and bias on each neuron, have also to be learned to achieve the function of classification or other tasks. Back-propagation is the common method that trains multi-layer neural networks. In multi-layer network, the errors can be back-propagated layer by layer and then the gradient of parameters can be obtained by the back-propagated errors. However, in NNN, there are no regular layers which leads to irregular depth. Even though the errors can also back-propagate from output layer, the different depth leads to different error decay. Then, the different error decay leads to uneven impact of reference output on input neurons. Therefore, some weights may not be well trained by directly using back-propagation.
\subsection{NNN's Parameter Learning Model}
In the parameter learning process, we should not only consider the errors in the output layer, but also consider the representation capability for input data. As a consequence, similar to RBM, we construct a probability model with the errors in the output layer as the energy:
\begin{equation}
\label{eq:proba}
p(x|y;\theta)=\frac{\exp(-E(x,y;\theta))}{\sum_{\chi}\exp(-E(\chi,y;\theta))}
\end{equation}
where $E(x,y;\theta)$ is the square error between output of the network $f_{\theta}(x)$ and reference output $y$, i.e., $E(x;\theta)=\|y-f_{\theta}(x)\|_2^2$ with $\theta$ being the network parameter set. Similar to the architecture model, $\chi$ denotes all the possible data in the whole data space. This model denotes a parameterized probability density that captures the distribution of observed data given a reference output.

\subsection{Optimization}
This model can be solved via the gradient of log-likelihood:
\begin{equation}
\label{eq:log}
\begin{aligned}
\triangle\theta=&\frac{\partial \log p(x|y;\theta)}{\partial \theta}\\
=&\frac{\partial -E(x)}{\partial \theta}-\sum_{\chi}\frac{\exp(-E(\chi))}{\sum_{\chi}\exp(-E(\chi))}\frac{\partial-E(\chi)}{\partial \theta}\\
=&\frac{\partial -E(x)}{\partial \theta}-\mathds{E}_{p(\chi|y;\theta)}\frac{\partial -E(\chi)}{\partial \theta}
\end{aligned}
\end{equation}
where $\mathds{E}$ denotes the expectation. The first term in Eq. \eqref{eq:log} is easy to compute with the back-propagation algorithm. For the second term, we use the data $\hat{x}$ sampled from the distribution $p(x|y;\theta)$ to estimate the expectation of gradients. However, with the complex architecture, it is difficult and even impossible to compute the probability of each input neuron. Therefore, we propose to solve the problem in reverse.

The sampled data is more likely to locate in the high density region of the distribution. Therefore, we can drive a random sample to move to the high density region by gradient, i.e., $\triangle\hat{x}=\partial p(\hat{x}|y;\theta)/\partial \hat{x}$. During the sampling process, the network parameter set $\theta$ is fixed which leads to the dominator in Eq. \eqref{eq:proba} being a constant. Then the updating gradient is computed as follows:
\begin{equation}
\label{eq:sample}
\triangle\hat{x}=\frac{\partial\exp(-E(\hat{x}))}{\partial\hat{x}}=\exp(-E(\hat{x}))\frac{\partial-E(\hat{x})}{\partial \hat{x}}
\end{equation}
The gradient is then the error back-propagated from the output layer. The sampled data $\hat{x}$ is then obtained by the gradient iteratively. This process will repeat several times to obtain several sampled data and estimate the expectation. After that, the gradient in Eq. \eqref{eq:log} can be computed by using back-propagation for each term.
%
%

This new model and learning process can relieve the uneven impact of output neurons. The updating gradient is the difference between gradients of observed data and sampled data. For a deep path, there is more error decay from the output layer. Then the components of deep paths in a sampled data is more random. Thus the gradient difference between observed data and sampled data will be larger. While for a shallow path, the gradient difference will be smaller. Then the uneven impact will be offset by the uneven difference.

\subsection{Toy Example and Potential Application}
\begin{figure}
  \centering
  \includegraphics[width=0.9\linewidth]{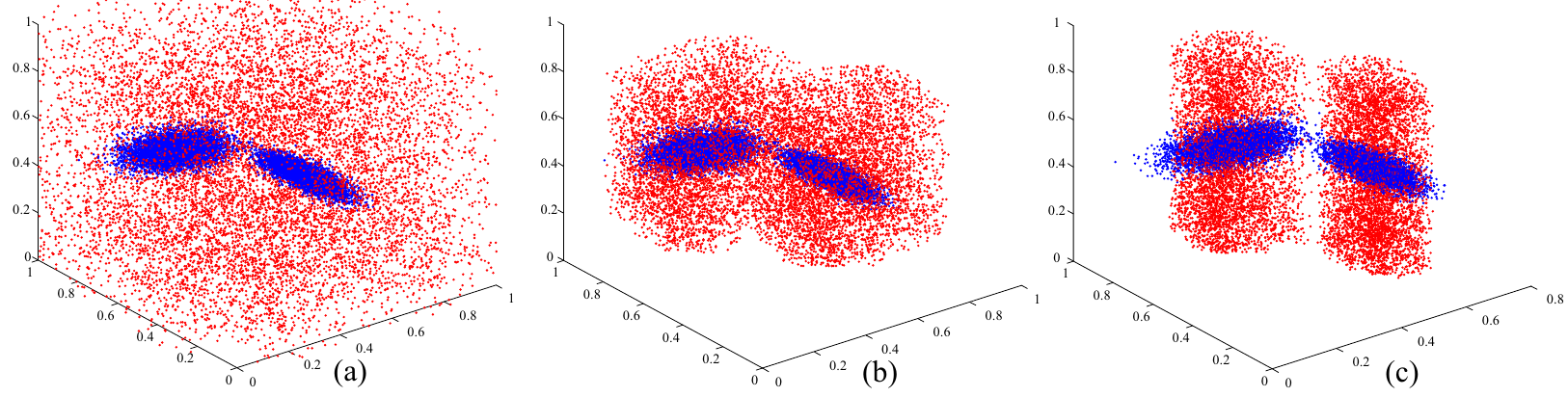}\\
  \caption{Distribution of observed data and sampled data in different learning stages. The blue points denote the observed data and red ones are sampled data. (a) Sampled distribution by randomly initialized network. (b) Sampled distribution by the network during the learning process. (c) Sampled distribution by the network after training.}\label{fig:toyexampleparameter}
  \vspace{-10pt}
\end{figure}

Similarly, we provide a toy example to explain the observed distribution and sampled distribution by the network. We train the network architecture and parameters by a simulated data set. There are 3 attributes in each data. The first two attributes follow the joint Gaussian distribution and the last attribute is a constant. The distribution of the simulated data is shown by the blue points in Fig. \ref{fig:toyexampleparameter}. There are two classes and different classes follow Gaussian distribution with different parameters.

In Fig. \ref{fig:toyexampleparameter}, the distribution assigned by the network is shown by the red points. With the randomly initialized network, the sampled data $\hat{x}$ are randomly distributed. During the learning process, the network begins to capture the distribution of observed data.
After learning, the learned network can well capture the distribution of the first two attributes in the observed data. Because the first two attributes play important roles in decision, the network assigns more connecting density for them. Then the network can learn to follow the distribution of them by connecting weights and biases. The last dimension has no contributions to classification and thus that of the sampled data is approximately randomly distributed after learning.

This novel phenomenon demonstrates NNN is robust to irrelevant components, i.e., the background. The sampled distribution represents the energy driven response of the network to different data. That means, the samples at the higher density region have lower energy, i.e., output difference from reference. Therefore, NNN assigns high response to the data indicated by red points in Fig. \ref{fig:toyexampleparameter}. In practical, there are more complex data and some attributes are not always background. But with the generative model derived objective function, the connecting density and the learned network parameters are expected to well capture the distribution of input data.
As a consequence, in this paper, we construct a novel dataset to test NNN.

\section{Reconstructed Dataset and Experiments}
In supervised learning, it is usually supposed that the training data set and test data set are independent and identically distributed. Therefore, large scale labeled dataset coving most cases in practical applications is necessary to train a robust model. Transfer learning \cite{transfer1} poses this problem and solves it by transferring the knowledge from labeled data in source domain to unlabeled data in target domain.
But transfer learning methods suppose that source domain and target domain are not independent.
From the toy example, NNN is able to generate high response to unseen patterns which shows potential significant applications. Therefore a new dataset is constructed based on the MNIST dataset to verify the superiority of NNN over traditional learners.

\subsection{Data Set}

\begin{wrapfigure}{r}{0.5\linewidth}
  \centering
  \includegraphics[width=1\linewidth]{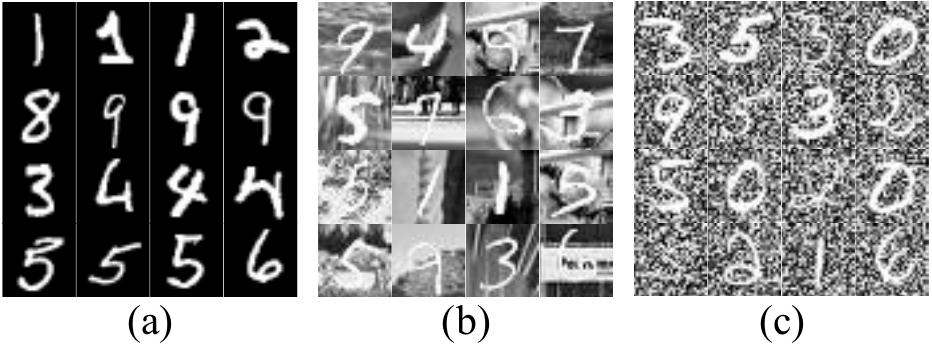}\\
  \caption{Some images in training and test data sets. (a) Training set with pure background. (b) Test set with random image patches as background. (c) Test set with random noise as background.}\label{fig:dataset}
  \vspace{-10pt}
\end{wrapfigure}
In the new dataset, we train the learners by the training set in the MNIST data set and apply the trained learner to digits with various backgrounds. Fig. \ref{fig:dataset} shows some images in the training set and test set. MNIST data set is a collection of handwritten digits (0-9) which is used to train and test classifiers for handwritten digit recognition. However, in practice, there are also digits on various backgrounds. Then some derivatives of MNIST data set are created, including MNIST digits with random image background (bg-img) and random noise background (bg-rand) \cite{DAE}. In super robust learning, the classifiers should recognize such digits when trained by pure digits.

In the experiments, 60000 training digits in the MNIST data set are used to train the classifiers. Then 50000 digits with bg-img and 50000 digits with bg-rand are used to test the classifiers. It deserves to be noted that training and test processes are independent, i.e., the digits in test set have no influence on the training process. To my knowledge, few methods could perfectly solve this problem. We set the number of neurons in the nucleus to be 200 and $\lambda=1$. We compare the proposed NNN with architectures of DBN and CNN. We set the scale of DBN as ``784-500-300-10'' \cite{DBN1} and CNN as LeNet in \cite{CNN1}. All the three architectures are trained by both back-propagation (BP) and the proposed error driven probability model (EDPM). The experimental results are evaluated by receiver operating characteristic (ROC) curve, precision-recall (PR) curve, area under ROC curve (AUC), area under PR curve (AP), and the classification accuracy (CA).

\subsection{Experimental Results}

\begin{wrapfigure}{r}{0.5\linewidth}
  \centering
  \includegraphics[width=\linewidth]{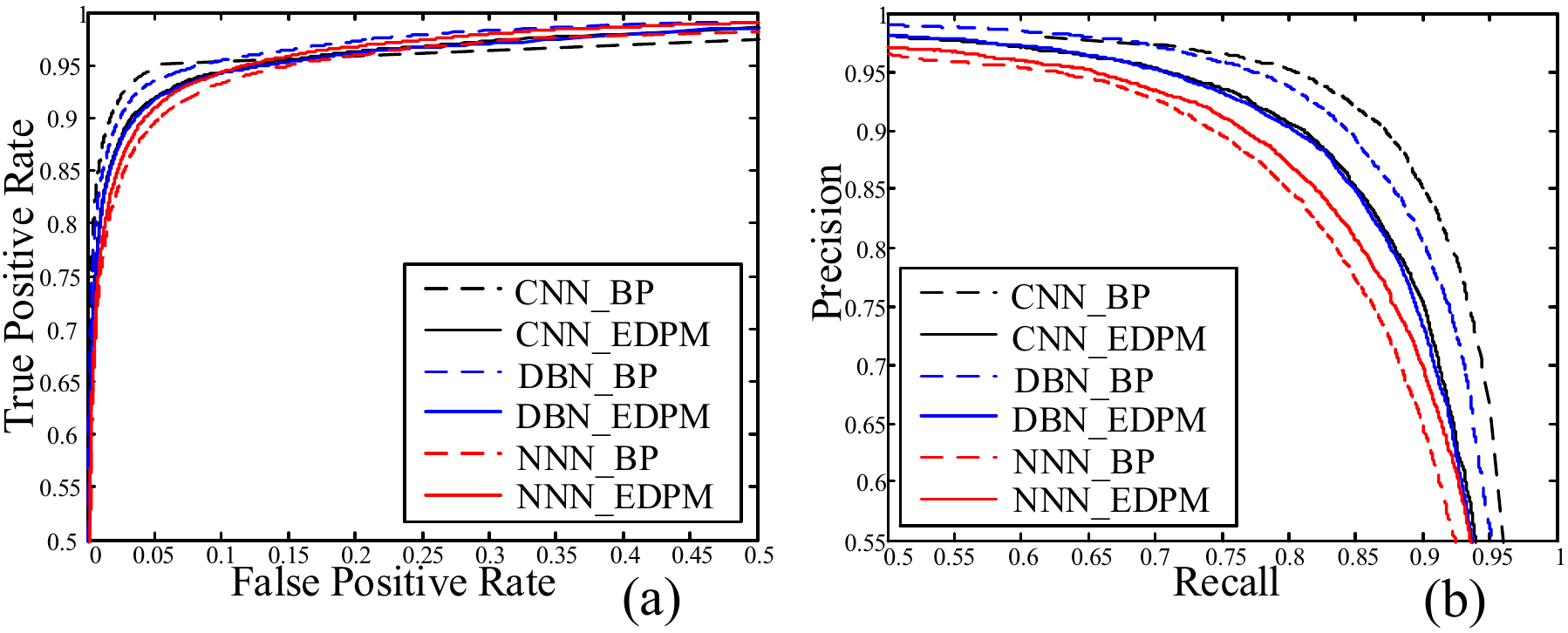}\\
  \caption{ROC and PR curves of the test result on digits in MNIST test set. (a) ROC curves (b) PR curves}\label{fig:mnistcurves}
\end{wrapfigure}
First we demonstrate the learning capability of NNN on traditional classification problem where the digits in training and test set follow the same distribution. After trained by the training set in MNIST data set, the classifiers are used to classify the digits in the test set. The ROC and PR curves of each classifier is shown in Fig. \ref{fig:mnistcurves}. It can be found that NNN cannot outperform traditional classifiers in this problem. The values of AUC, AP, and CA are listed in Table \ref{tab:mnistcurves}. CNN achieves the best performance from the classification accuracy due to its special architecture. Although NNN achieves the lowest classification accuracy, the accuracy is close to that of DBN. It deserves to be noted that there are only 200 neurons in the nucleus and 146686 connections in the whole network after training. In DBN, there are 800 hidden units and 545000 connections and much more in CNN. Therefore, NNN achieves equivalent performance to that of DBN with less processing units and parameters.

The superiority of NNN is its robustness to irrelevant components in input data, i.e., background in images. Therefore, we continue to use the trained classifiers to classify the digits with various backgrounds.
\begin{figure}
  \centering
  \includegraphics[width=\linewidth]{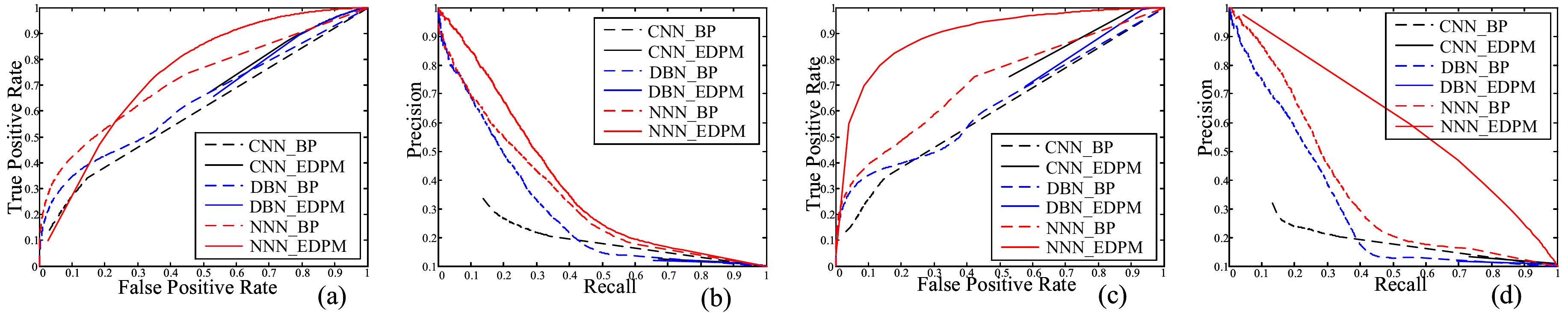}\\
  \caption{ROC and PR curves of the test result on digits with currupted background. (a) ROC curves on test data with bg-img (b) PR curves on test data with bg-img (c) ROC curves on test data with bg-rand (d) PR curves on test data with bg-rand}\label{fig:bgcurves}
  \vspace{-10pt}
\end{figure}

The ROC and PR curves of results on the two test sets, i.e., MNIST digits with bg-img and bg-rand, are shown in Fig. \ref{fig:bgcurves} respectively. For the digits with bg-img, the curves of NNN cover most of other curves and for the digits with bg-rand, the curves of NNN cover all the other curves. Traditional architectures are exhausted to deal with such a problem. Because they are composed of hierarchical layers, input neurons have approximately equal impact on the output neurons from architecture. During the back-propagation process, the output errors will not influence the weights of background pixels a lot due to the weak contribution of them. Therefore, back-propagation could not weaken the influence of background changes on output.
In NNN, the less relevant neurons will be directed into deeper and narrower paths because they have much less impact on the output neurons. Therefore, NNN is more robust and greatly outperforms the traditional architectures.

Then the AUC, AP, and AC values of the two test sets are listed in Table \ref{tab:mnistcurves} respectively. It can be found that CNN achieves worst performance in this problem. Because in CNN, the parameters in local convolutional kernels are shared, the background and foreground use the same kernels. Therefore, the influence of background neurons on the output neurons will be larger compared with the architectures of DBN and NNN. 
The digits with bg-img are more difficult to recognize because the background greatly disturb the digits as shown in Fig. \ref{fig:dataset}. Although NNN cannot achieve equivalent performance compared with traditional classifiers trained by images with the same kind of background as shown in \cite{DAE}, it outperforms traditional classifiers greatly when the backgrounds of training and test sets are different. The robust learning and inference capability of NNN is larger than that of traditional learning architectures in this data set.



\begin{table}[h]
\centering
\small
\caption{AUC, AP, and CA values of the test result on digits in different test set. Arc. denotes architecture and Par. denotes parameter learning method.}
\begin{tabular}{c||c|c|c|c|c|c|c}
\toprule
\multirow{2}{*}{Test Set}&Arc.&\multicolumn{2}{c|}{DBN}&\multicolumn{2}{c|}{CNN}&\multicolumn{2}{c}{NNN}\\ \cline{2-8}
&Par.&BP&EDPM&BP&EDPM&BP&EDPM\\ \Xhline{0.7pt}
\multirow{3}{*}{MNIST}&AUC&\textbf{.9968}&.9888&.9864&.9898&.9846&.9921\\ 
&AP&\textbf{.9941}&.9896&.3561&.9921&.8734&.9814\\ 
&CA(\%)&98.97&98.74&\textbf{99.14}&98.73&98.69&98.86\\ \hline

\multirow{3}{*}{bg-img}&AUC&.6395&.4005&.6002&.4139&.7205&\textbf{.7434}\\ 
&AP&.2860&.0389&.1514&.0379&.3300&\textbf{.3776}\\ 
&CA(\%)&29.71&10.28&24.24&13.14&37.18&\textbf{57.33}\\ \hline

\multirow{3}{*}{bg-rand}&AUC&.6230&.3671&.5981&.4183&.7082&\textbf{.8926}\\ 
&AP&.3020&.0340&.1478&.0323&.3658&\textbf{.5685}\\ 
&CA(\%)&30.66&9.55&22.18&15.62&32.84&\textbf{67.00}\\ \bottomrule
\end{tabular}
\label{tab:mnistcurves}
\vspace{-10pt}
\end{table}

\section{Conclusions and Future Work}
This paper attempts a new neural network architecture which is called nucleus neural network (NNN). NNN mimics nuclei in brain where there is no regular layers so as to relax the limitation of layers. To evaluate the architectures efficiently, we directly model the architecture without involvement of connecting weights and biases.
After optimization of architecture and parameters respectively, we find that this novel architecture is robust to irrelevant components in data.
Therefore, we reconstruct a new dataset based on the MNIST dataset. From the experiments, NNN is superior over traditional deep learners on this data set.

However, there are still many defects including the low architecture and parameter learning efficiency and unsatisfactory classification accuracy. In the future work, we will further improve the efficiency of learning methods and attempt feedback connections. Moreover, we will stack NNN to mimic the hierarchical architecture in brain.


\bibliographystyle{IEEEtran}
\bibliography{references}

\end{document}